\documentclass{article}

\usepackage[preprint,nonatbib]{neurips_2024}

\usepackage[utf8]{inputenc} 
\usepackage[T1]{fontenc}    
\usepackage[pagebackref,breaklinks,colorlinks]{hyperref}       
\usepackage{url}            
\usepackage{booktabs}       
\usepackage{amsfonts}       
\usepackage{nicefrac}       
\usepackage{microtype}      
\usepackage{xcolor}         

\usepackage[pdftex]{graphicx}
\usepackage{amsmath}
\usepackage{multirow}
\usepackage{caption}
\usepackage{subcaption}
\usepackage{amssymb}
\usepackage{makecell}
\usepackage{changes}
\usepackage{wrapfig}
\usepackage{color}
\usepackage{colortbl}
\usepackage{bm}

\title{Aligning in a Compact Space: Contrastive Knowledge Distillation between Heterogeneous Architectures}

\author{%
  \makecell*[c]{Hongjun~Wu$^{1}$, Li~Xiao$^{1}$\thanks{Corresponding author.}, Xingkuo~Zhang$^{1}$, Yining~Miao$^{1}$} \\
  $^1$School of Artificial Intelligence, Beijing University of Posts and Telecommunications. \\
  \makecell*[c]{\texttt{\{hongjun.wu, andrewxiao, dd\_zxk, yiningmiao\}@bupt.edu.cn}}
}

\begin{document}

\maketitle

\begin{abstract}
\label{sec:abstract}
Knowledge distillation is commonly employed to compress neural networks, reducing the inference costs and memory footprint. In the scenario of homogenous architecture, feature-based methods have been widely validated for their effectiveness. However, in scenarios where the teacher and student models are of heterogeneous architectures, the inherent differences in feature representation significantly degrade the performance of these methods. Recent studies have highlighted that low-frequency components constitute the majority of image features. Motivated by this, we propose a Low-Frequency Components-based Contrastive Knowledge Distillation (LFCC) framework that significantly enhances the performance of feature-based distillation between heterogeneous architectures. Specifically, we designe a set of multi-scale low-pass filters to extract the low-frequency components of intermediate features from both the teacher and student models, aligning them in a compact space to overcome architectural disparities. Moreover, leveraging the intrinsic pairing characteristic of the teacher-student framework, we design an innovative sample-level contrastive learning framework that adeptly restructures the constraints of within-sample feature similarity and between-sample feature divergence into a contrastive learning task. This strategy enables the student model to capitalize on intra-sample feature congruence while simultaneously enhancing the discrimination of features among disparate samples. Consequently, our LFCC framework accurately \textbf{captures the commonalities in feature representation} across heterogeneous architectures. Extensive evaluations and empirical analyses across three architectures (CNNs, Transformers, and MLPs) demonstrate that LFCC achieves superior performance on the challenging benchmarks of ImageNet-1K and CIFAR-100. All codes will be publicly available.
\end{abstract}

\section{Introduction}
\label{sec:intro}
Knowledge distillation has emerged as an extensively utilized strategy for model compression. It typically employs a teacher-student architecture, where the student model is guided to mimic the teacher's predictions\cite{distilling2015hinton, decoupled2022zhao, knowledge2022huang, ofa2023hao} or feature representations\cite{fitnets2015Adriana, correlation2019peng, relational2019park, contrastive2019tian}, thereby enhancing the performance of compact models. Current methods predominantly focus on scenarios where both the teacher and student models share the same architectural lineage, often involving pairing a larger-scale teacher model with a smaller-scale student, exemplified by transitions such as ResNet18 to ResNet34\cite{deep2016he} or ViT-S to ViT-G\cite{image2020dosovitskiy}.

However, the performance ceilings of models vary due to the differing scalability of the architectures. This limitation narrows the applicability of knowledge distillation homogeneous architectures. For instance, it is challenging for a student model from the MobileNet\cite{mobilenets2017howard} family to find a teacher model within its own lineage that can match the capabilities of ViT-G\cite{image2020dosovitskiy}. Knowledge distillation in heterogeneous architectures addresses this limitation by expanding the pool of potential teacher-student pairings. While some research has ventured into exploring teacher-student combinations in heterogeneous architectures\cite{knowledge2019luo, customizing2019shen, heterogeneous2020passalis, ofa2023hao, promoting2024zheng}, the field is still in its early stages and holds substantial potential for further exploration and advancement.

Existing knowledge distillation methods primarily employ logit\cite{distilling2015hinton, decoupled2022zhao, knowledge2022huang, ofa2023hao} or intermediate feature maps\cite{fitnets2015Adriana, correlation2019peng, relational2019park, contrastive2019tian} as the medium for knowledge transfer. The logit space serves as a task-specific shared domain, applicable to all knowledge distillation tasks. In contrast to logit-based approaches, feature-based methods transmit knowledge via aligning intermediate feature representations point-to-point. While these methods often excel in knowledge distillation with homogeneous architectures due to the intrinsic similarity of feature representations, they face serious setback in heterogeneous settings\cite{ofa2023hao}. The substantial differences in meta-architecture, input format, and spatial representation between heterogeneous models preclude the use of simple similarity metrics for aligning stage-wise feature representations. Moreover, mimicing the local details in feature representations may lead to detrimental guidance. Therefore, extracting effective common knowledge from the potential feature space of heterogeneous models remains a pressing issue.

Inspired by the effectiveness of low-frequency components in feature representation\cite{vtc2022wang, head2023dong}, we propose a Low-Frequency Components-based Contrastive Knowledge Distillation (LFCC) framework for knowledge distillation in heterogeneous architectures. The overarching strategy involves extracting and condensing the low-frequency components of image features into an aligned compact space. Specifically, we employ a multi-scale low-pass filter, and its learnable derivative, to extract and compress the low-frequency components of both teacher and student features into a compact categorical representation. This approach mitigates the substantial representational disparity between heterogeneous models, preventing detrimental guidance from spatial noise. Harnessing the intrinsic pairing of the teacher-student framework, we design a novel sample-level contrastive learning framework that transforms the feature-based knowledge distillation into a sample-level contrastive learning. This framework treats the representations of the student and teacher for the same sample as positive pairs and contrasts them with other samples in the same batch as negative pairs. By mapping positive pairs closer together and negative pairs further apart, the student model not only benefits from the intra-sample feature similarity but also extracts hidden knowledge from inter-sample feature disparity. To entirely evaluate the proposed LFCC framework, we conduct extensive experiments across 15 model combinations involving three heterogeneous architectures. Our method surpasses the prior state-of-the-art OFA-KD\cite{ofa2023hao} approach, achieving up to a 0.83\% and 4.19\% improvement on ImageNet-1K and CIFAR-100, respectively.  Moreover, extensive and thorough ablation experiments validate the effectiveness of the LFCC.

\section{Related works}
\label{sec:related_works}

\subsection{Knowledge distillation}
The recent methods for knowledge distillation can be roughly categorized into two aspects, \textit{i.e.}, logits-based methods and feature-based methods.

\textbf{Logits-based knowledge distillation}\cite{li2022knowledge, zhang2023not, mirzadeh2020improved} enhances a smaller student model by transferring soft targets from a larger teacher model. Hinton et al.\cite{distilling2015hinton} introduce a method to soften logits in the softmax function, guiding the students to acquire knowledge from the teacher's prediction. Zhao et al.\cite{decoupled2022zhao} decomposes the logit-based distillation loss into the target domain and the non-target domain, so as to learn the dark knowledge in the non-target domain. Recognizing the challenge for students to learn from stronger teachers, Huang et al.\cite{knowledge2022huang} suggest a relation-based loss to mitigate the strict match requirement of KL divergence. Sun et al.\cite{Logit2024Sun} highlight issues with a shared temperature in KD and propose logit normalization to allow students to focus more on the teacher's logits internal relations. Hao et al.\cite{ofa2023hao} present a KD framework for heterogeneous architectures, employing projectors to align student features with teacher outputs. 

\textbf{Feature-based knowledge distillation}\cite{huang2023pixel, mobile_sam, lin2022knowledge} involves transferring knowledge within the feature space. FitNet\cite{fitnets2015Adriana} introduces a method that utilizes features from intermediate layers of a teacher model to guide the learning process of the student model. Unlike conventional point-to-point knowledge transfer methods, RKD\cite{relational2019park} proposes learning the relations between output feature maps. CRD\cite{contrastive2019tian} employs a contrastive learning approach to maximize the lower bound of mutual information between teacher and student. FSPNet\cite{gift2017yim} contends that the variations in features between each layer represent learnable knowledge. CC\cite{correlation2019peng}, also based on relational knowledge distillation, proposes a generalized kernel method based on Taylor series expansion to better capture relationships between features. CAT-KD\cite{guo2023class} propose a highly interpretable knowledge distillation method based on class activation map. Yang et al.\cite{masked2022yang} believes that directly mimicking the teacher is not necessary to enhance the expressive power of student features. Instead, they propose MGD which initially masks student features randomly and then employs a simple block to generate complete teacher features from the masked features. FGD\cite{yang2022focal} enhances the effectiveness of knowledge distillation in object detection by employing focal distillation and global distillation.

\subsection{Computer vision models}

To comprehensively evaluate the generalization of our approach, we select teacher-student combinations from three meta-architectures, including CNNs, Transformers, and MLPs.

\textbf{Convolutional Neural Networks} (CNNs) are adopted at image processing due to their inherent inductive biases, translational invariance, and the sliding window mechanism. Researchers have introduced several enhancements to CNNs, including residual connections\cite{deep2016he}, depthwise separable convolutions\cite{mobilenets2017howard}, systematic model scaling\cite{efficientnet2019tan}, deep neural architecture search\cite{radosavovic2020designing}, and the use of large convolutional kernels\cite{convnet2022liu}. These innovations have reinforced CNNs' pivotal role in the field of image processing, attributable to their superior performance and robust generalization capabilities.

\textbf{Transformers}, initially designed for Natural Language Processing (NLP) tasks\cite{vaswani2017attention}, have increasingly been integrated into computer vision. The introduction of Vision Transformers (ViT)\cite{image2020dosovitskiy} has catalyzed a significant convergence between language and vision tasks, revolutionizing traditional network architecture design. ViT processes images by segmenting them into a series of patches for sequential input, outperforming classic CNN models. Subsequent advancements, such as window attention\cite{swin2021liu} and pyramidal architectures\cite{zhanghivit}, have further expanded the capability of Transformer family. Additionally, Transformers have been extensively applied in downstream vision tasks, such as object detection\cite{end2020carion}, semantic segmentation\cite{segformer2021xie} and image generation\cite{peebles2023scalable}.

\textbf{Multi-Layer Perceptrons} (MLPs), integral to the success of Transformers, have garnered increased research attention. Given that MLPs are simple nonlinear mappings that are less sensitive to local information, strategies are essential to adapt them for image processing. These strategies include token or channel mixing\cite{mlp2021tolstikhin}, affine transformations\cite{resmlp2022touvron}, gating mechanisms\cite{pay2021liu}, structural reparameterization\cite{repmlpnet2022ding}, and spatial splitting\cite{hou2022vision}.

\section{Method}
\label{sec:method}
\subsection{Recap of knowledge distillation}

Knowledge distillation aims to transfer knowledge from a large-size teacher model \textbf{T} to a small-size student model \textbf{S}. This process often uses logits and features as carriers for knowledge transfer. 

\paragraph{Logits-based Distillation.}
Logits carry category distribution information that directly impacts classification probabilities. Logit-based approach has proven to be an effective strategy to circumvent the discrepancies in the spatial distribution of features among heterogeneous models, as recent research has demonstrated. Specifically, the output category distribution of the student model can be made to approximate that of the teacher model, thereby improving the performance of the student model. 

\begin{equation}
\label{eq:KD_loss}
    \mathcal{L} = \mathcal{L}_{\text{CE}}(\mathbf{p}^{s}, y) + \lambda \mathcal{L}_{\text{KLD}}(\mathbf{p}^{s}, \mathbf{p}^{t}),
\end{equation}
where \(p^s\) and \(p^t\) are the predicted logits of student model and teacher model, respectively. \(\mathcal{L}_{\text{KLD}}\) is the Kullback-Leibler divergence loss function. \(y\) is the one-hot groundtruth labels. \(\lambda\) is the hyperparameter adjusts the weight of soft label \(p^t\). 

\paragraph{Feature-based Distillation.}

By minimizing the Euclidean distance between features, feature distillation offers a finer-grained supervision for the student model through the introduction of sample-level comparisons. Formally, the objective of feature distillation approaches can be written as:

\begin{equation}
    \mathcal{L} = \mathcal{L}_{\text{CE}}(\mathbf{p}^{s}, y) + \lambda_{\text{KLD}} \mathcal{L}_{\text{KLD}}(\mathbf{p}^{s}, \mathbf{p}^{t}) + \lambda_{\text{MSE}} \sum_{i=1}||\mathrm{F}^{t}_{i}-\phi (\mathrm{F}^{s}_{i})||^2,
\end{equation}
where \(F^s\) and \(F^t\) are the features extracted from the i-th layer of student model and teacher model. \(\phi(\cdot)\) is the projector that transform student feature to match the dimension of teacher feature. \(\lambda_{\text{KLD}}\) and \(\lambda_{\text{MSE}}\) balance the logit loss and feature loss.

Existing feature distillation methods simply align feature dimensions through convolution or MLP modules, which is effective when both the teacher and student models belong to the same meta-architecture. However, models with different meta-architectures exhibit significant differences in terms of model inputs, inductive biases, and spatial distributions, among others. Consequently, enforcing complete feature-level alignment between the student and teacher models may not be the optimal approach.

\begin{figure}
  \centering
  \includegraphics[width=\linewidth]{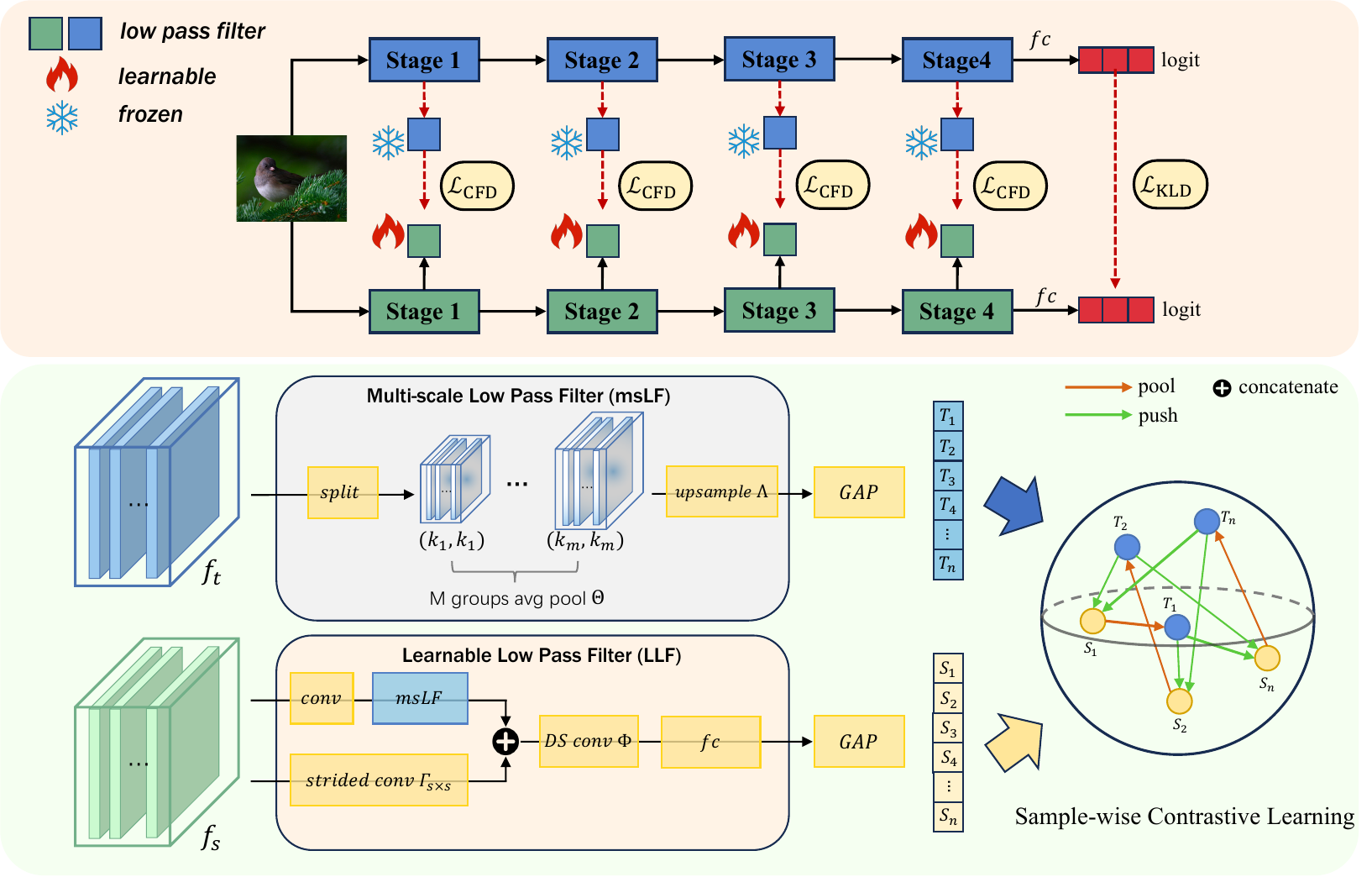}
  \caption{The pipeline of our LFCC framework. The overall architecture of the LFCC framework is located in the top half, where we align the features output from each stage via low-pass filters and contrastive learning. The detailed process of the alignment module is located in the lower part.}
  \label{fig:pipeline}
  \vspace{-0.5cm}
\end{figure}

\subsection{Low-Frequency Components-based Contrastive Knowledge Distillation}
\label{subsec:method}
We propose a novel feature-based distillation framework, named Low-Frequency Components-based Contrastive Knowledge Distillation (LFCC). Figure \ref{fig:pipeline} provides an overview of the proposed LFCC.

\paragraph{Multi-scale Low-Pass Filters.}
\label{para:msLF}

In light of the substantial discrepancies in feature representation between heterogeneous teacher-student models, we propose to align features from a frequency domain perspective. Low-frequency components account for the majority of the content in an image and convey most of the semantic information\cite{vtc2022wang}, while high-frequency information often contains noise. Consequently, we employ low-pass filters to extract the low-frequency components from image features. 

We devise two low-pass filters tailored for the teacher and student branches, each with differing levels of learnability. For the teacher branch, we adopte a conventional average pooling as the low-pass filter and craft multi-scale filters by adjusting different kernel sizes and strides across multiple groups to accommodate varying cutoff frequencies. For the \(m\)-th group, we have

\begin{equation}
\text{msLF}_m(f_m) = \Lambda(\Theta_{k\times k}(f_m))
\end{equation}

where \(\Theta_{k\times k}\) denotes the average pooling function with the output size of \(k\times k\). \(\Lambda(\cdot)\) is the bilinear interpolation to align the size of each group.

Meanwhile, for the student branch, we engineer a learnable low-pass filter (LLF) composed of a multi-scale low-pass filter, a convolutional downsampling module, and a depth-wise separable convolution (DSConv). The learnable low-pass filter can be represent as

\begin{equation}
    \text{LLF}(f) = fc(\Phi_{3\times 3}([\Gamma_{s\times s}(f)), \text{msLF}(f)])
\end{equation}

where \(\Phi_{3\times 3}\) is a depth-wise separable convolution with kernel size \(3\times 3\). \(\Gamma_{s\times s}\) represents a convolution layer with kernel size 
\(s\times s\) and stride \(s\times s\). \([\cdot]\) denotes concatenate operation. \(fc\) aligns the number of channels by linear projection. \(f\) denotes the intermediate feature.

\paragraph{Aligning features in a compact space.}
\label{para:compat}

The models with heterogeneous architectures exhibit significant differences in feature representation, which can lead to the introduction of detrimental spatial noise through direct feature alignment, thereby degrading the performance of the student model.  We observe that mainstream image classification models typically map the results of global average pooling or a preset cls embedding into the logit space during the final stage, indicating that the cls embedding encapsulates sufficient semantic information to describe the image.  In our framework, we use global feature representation to measure the differences between the features of the two branches and achieve feature alignment within a more compact space, thus avoiding the noise introduced from the local details of the representation and reducing the difficulty of feature alignment in heterogeneous settings.

\paragraph{Sample-level constrastive learning.}
\label{para:constra}

Traditional feature distillation methods focus on guiding the student model to learn the teacher model's feature representation via positive sample comparison, overlooking discrimination between positive and negative samples. The teacher-student framework, where sample pairs are formed by the outputs of the teacher and student models, renders contrastive learning suitable for knowledge distillation. We introduces sample-level contrastive learning in the feature space, enabling fine-grained learning by comparing the features at each stage of the teacher and student models.

For feature distillation, contrastive learning can be formulated as the distance between the features of teacher and student:

\begin{equation}
\label{eq:dist1}
    \mathcal{L}_{\text{CFD}} = \mathbb{E}_{\bm{x_i}\sim \mathcal{X}} \left[ \text{dist}(t(x_i), s(x_i)) - \mathbb{E}_{\bm{x_{j}}\sim \mathcal{X}} \text{dist}(t(x_i), s(x_j)) \right]
\end{equation}

where \(\mathcal{X}\) is the distribution of samples. \(t(\cdot)\) and \(s(\cdot)\) represent the teacher model and student model. Hence, teacher features \(t(x_i)\) and student features \(s(x_i)\) constitute a positive pair, while \(t(x_i)\) and \(s(x_j)\) form a negative pair. We set the distance metric as \(\mathcal{D}=-dist(a, b)\) to represent the similarity between \(a\) and \(b\). The formula \ref{eq:dist1} can be rewritten as:

\begin{equation}
    \begin{aligned}
        \mathcal{L}_{\text{CFD}} = & \mathbb{E}_{\bm{x_i}\sim \mathcal{X}} \left[ -(\mathcal{D}(t(x_i), s(x_i)) -  \mathbb{E}_{\bm{x_{j}}\sim \mathcal{X}} \mathcal{D}(t(x_i), s(x_j)) \right] \\
        \simeq & \mathbb{E}_{\bm{x_i}\sim \mathcal{X}} -log \left[ \frac{\exp(\mathcal{D}(t(x_i), s(x_i)))}{\mathbb{E}_{\bm{x_{j}}\sim \mathcal{X}} \exp(\mathcal{D}(t(x_i), s(x_j)))}  \right] \\
    \end{aligned}
\end{equation}

To facilitate computation, we confine the sample space for contrastive learning to a single batch \(B\). The goal is to learn with \(t(x_i)\) as the anchor, distinguishing the positive sample \(s(x_i), i=1,2,\cdots,B\) from other student features \(s(x_j), j=1,2,\cdots ,i-1,i+1,\cdots , B\) within the batch. Here we use cosine similarity as a distance metric, and at the \(l\)-th stage, the loss function can be expressed as:

\begin{equation}
    \mathcal{L}_{\text{CFD}}(l) = -\frac{1}{B} \sum^B_{i=1} \mathbb{E}_{\bm{x_i}\sim \mathcal{X}} log \left[ \frac{\exp(t_l(x_i)\cdot s_l(x_i) / \tau )}{ \sum_{j, j\neq i}  \mathbb{E}_{\bm{x_{j}}\sim \mathcal{X}}  \exp((t_l(x_i)\cdot s_l(x_j) / \tau))}  \right],
\end{equation}

\begin{table}
  \centering
  \small
  \renewcommand\tabcolsep{2.7pt}
  \renewcommand{\arraystretch}{0.96}
  \caption{Comparison to state of the art KD methods with heterogeneous architectures on ImageNet-1K. \underline{Underlined}: the second best result. \textbf{Bold}: the best result. $*$: results achieved by combining with FitNet\cite{fitnets2015Adriana}.}
  \begin{tabular}{cc cc cccc cccc>{\columncolor[gray]{0.85}}c}
    \toprule
    \multirow{2}{*}{Teacher} & \multirow{2}{*}{Student} & \multicolumn{2}{c}{From Scratch} & \multicolumn{4}{c}{Logits-based} & \multicolumn{5}{c}{Feature-based}                                                                                                                                             \\
    \cmidrule(lr){3-4} \cmidrule(lr){5-8} \cmidrule(lr){9-13}
                             &                          & T.                               & S.                               & KD                           & DKD     & DIST   & OFA   & FitNet               & CC                            & RKD                           & CRD              &  LFCC           \\
    \midrule
    \multicolumn{2}{c}{\emph{CNN-based students}}                                                                                                                                                                                                                                                            \\
    \midrule
    DeiT-T                   & ResNet-18            & 72.17                             & 69.75                           & 70.22             & 69.39                          & 70.64              & \underline{71.34}      & 70.44   & 69.77 & 69.47 & 69.25     & \bf{71.40}        \\
    Swin-T                   & ResNet-18            & 81.38                             & 69.75                           & 71.14             & 71.10                          & 70.91              & \underline{71.85}            & 71.18    & 70.07 & 68.89 & 69.09  & \bf{72.51}        \\
    Mixer-B/16               & ResNet-18            & 76.62                             & 69.75                           & 70.89             & 69.89                          & 70.66                          & \underline{71.38}           & 70.78  & 70.05 & 69.46 & 68.40    & \bf{71.39}  \\
    DeiT-T                   & MobileNetV2              & 72.17                             & 68.87                           & 70.87             & 70.14                          & 71.08              & \underline{71.39}           & 70.95 & 70.69 & 69.72 & 69.60  & \bf{71.73}    \\
    Swin-T                   & MobileNetV2              & 81.38                             & 68.87                           & 72.05             & 71.71                          & 71.76                          & \underline{72.32}         & 71.75 & 70.69 & 67.52 & 69.58  &\bf{72.84}      \\
    Mixer-B/16               & MobileNetV2              & 76.62                             & 68.87                           & 71.92             & 70.93                          & 71.74                          & \bf{72.12}            & 71.59                            & 70.79 & 69.86 & 68.89    & \underline{72.05}    \\
    \toprule
    \multicolumn{2}{c}{\emph{Transformer-based students}}                                                                                                                                                                                                                                                            \\
    \midrule
    ResNet-50                 & DeiT-T               & 80.38                             & 72.17                           & 75.10             & 75.60\rlap{$^*$} & 75.13\rlap{$^*$} & \bf{76.55}\rlap{$^*$} & 75.84                & 72.56 & 72.06 & 68.53  &  \underline{76.04}  \\
    ConvNeXt-T               & DeiT-T               & 82.05                             & 72.17                           & 74.00             & 73.95                          & 74.07              & \underline{74.41}            & 70.45                            & 73.12 & 71.47 & 69.18   & \bf{75.24}    \\
    Mixer-B/16               & DeiT-T               & 76.62                             & 72.17                           & 74.16             & 72.82                          & 74.22                          & \underline{74.46}            & 74.38                & 72.82 & 72.24 & 68.23   & \bf{74.57}    \\
    ResNet-50                 & Swin-N               & 80.38                             & 75.53                           & 77.58             & 78.23\rlap{$^*$} & 77.95\rlap{$^*$} & \underline{78.64}\rlap{$^*$}  & 78.33                & 76.05 & 75.90 & 73.90  & \bf{78.71}  \\
    ConvNeXt-T               & Swin-N               & 82.05                             & 75.53                          & 77.15             & 77.00                          &  77.25              & \underline{77.50}             & 74.81                       & 75.79 & 75.48 & 74.15   & \bf{77.85}    \\
    Mixer-B/16               & Swin-N               & 76.62                             & 75.53                           & 76.26             & 75.03                          & 76.54              & \underline{76.63}            & 76.17                         & 75.81 & 75.52 & 73.38  & \bf{76.91}    \\
    \toprule
    \multicolumn{2}{c}{\emph{MLP-based students}}                                                                                                                                                                                                                                                            \\
    \midrule
    ResNet-50                 & ResMLP-S12               & 80.38                             & 76.65                           & 77.41             & 78.23\rlap{$^*$} & 77.71\rlap{$^*$} & \underline{78.53}\rlap{$^*$}  & 78.13                & 76.21 & 75.45 & 73.23  & \bf{78.64}  \\
    ConvNeXt-T               & ResMLP-S12               & 82.05                             & 76.65                          & 76.84             & 77.23                          & 77.24              & \underline{77.53}             & 74.69              & 75.79 & 75.28 & 73.57  & \bf{77.59}      \\
    Swin-T                   & ResMLP-S12               & 81.38                             & 76.65                           & 76.67             & 76.99                          & 77.25            & \underline{77.31}            & 76.48                         & 76.15 & 75.10 & 73.40  & \bf{77.65}     \\
    \bottomrule
  \end{tabular}
  \label{tab:imagenet}
  \vspace{-0.3cm}
\end{table}

where \(\tau\) denotes temperature parameter that modulates the discrimination against negative samples. We empirically set \(\tau\) to 0.1 in all experiments. \(t_l(\cdot)\) and \(s_l(\cdot)\) represent the feature at the \(l\)-th stage. The overall Low-pass Contrastive Feature Distillation objective for the student extends the loss in eq.\ref{eq:KD_loss} with \(\mathcal{L}_{\text{CFD}}\) provides sample-level supervision.

\begin{equation}
    \mathcal{L} = \mathcal{L}_{\text{CE}}(\mathbf{p}^{s}, y) + \lambda_{\text{KLD}} \mathcal{L}_{\text{KLD}}(\mathbf{p}^{s}, \mathbf{p}^{t}) + \lambda_{\text{CFD}} \sum_{l}\mathcal{L}_{\text{CFD}}(l).
\end{equation}


\section{Experiment}
\label{sec:exper}

\subsection{Dataset and Implementation}
We conduct comprehensive evaluation of the proposed LFCC method. In this section, we briefly introduce the dataset and Implementation, and the detailed setup is provided in the supplementary material.

\paragraph{Dataset}

We evaluate the heterogeneous distillation method on ImageNet-1K\cite{deng2009imagenet} and CIFAR-100\cite{krizhevsky2009learning}. The ImageNet-1K dataset consists of 1000 categories. The training set has about 128 million images and the validation set contains 50k images. The CIFAR-100 dataset contains 100 classes with 600 images per class. The training and validation sets contain 50K and 10K images.

\paragraph{Model}

We quantitatively evaluate our LFCC method on two datasets and compare it with both logit-based and hint-based approaches. The former includes vanilla KD\cite{distilling2015hinton}, DKD\cite{decoupled2022zhao}, DIST\cite{knowledge2022huang}, OFA-KD\cite{ofa2023hao}, and the latter contains FitNet\cite{fitnets2015Adriana}, CC\cite{correlation2019peng}, RKD\cite{relational2019park}, and CRD\cite{contrastive2019tian}. Additionally, we select models from three architectures: ResNet\cite{deep2016he} and MobileNet V2\cite{sandler2018mobilenetv2} from the CNN family, DeiT\cite{training2021touvron}, ViT\cite{image2020dosovitskiy}, and Swin Transformer\cite{swin2021liu} from the Transformer family, and MLP-Mixer\cite{mlp2021tolstikhin} and ResMLP\cite{resmlp2022touvron} from the MLP family. The Swin Nano and Swin Pico\cite{swin2021liu, ofa2023hao} are smaller variants of the Swin Tiny.

For feature-based distillation, alignment of feature representations between teacher and student models at each stage is essential. For consistency in comparison, we adopt the configuration from OFA-KD. In pyramid-structured models like ResNet\cite{deep2016he}, we position the feature alignment module at the conclusion of each stage. Conversely, for models with uniform layer structures, such as ViT\cite{image2020dosovitskiy}, we insert the module at the quartiles of the overall layer count.

\paragraph{Implementation}

Images from both datasets are resized to 224x224 for training. Different optimizers are tailored for each type of student model based on empirical observations: the CNN family employs the SGD optimizer, while the Transformer and MLP families use the AdamW optimizer. Students from CNN family are trained for 100 epochs on the ImageNet-1K dataset, while other families are trained for 300 epochs. For CIFAR-100, all students are trained for 300 epochs. We report the top-1 accuracy average on 3 runs in all experiments.

\subsection{Performance on ImageNet-1K}

The comparative experiments on ImageNet-1K involve student models from three architectures, 15 heterogeneous teacher-student combinations, and 9 knowledge distillation methods including the proposed LFCC method. The results are reported in Table \ref{tab:imagenet}. Compare to previous distillation methods, our LFCC achieves state-of-the-art performance in most teacher-student combinations. Particularly, when the student model follows a CNN architecture, our approach outperforms other distillation methods in five combinations. When training students based on Transformer or MLP architectures, LFCC demonstrates comprehensive advantages, achieving performance gains ranging from 0.06\% to 0.83\% over the state-of-the-art method OFA-KD\cite{ofa2023hao}. Notably, methods mark with * in the table combine logit-based methods with FitNet\cite{fitnets2015Adriana}, leveraging the advantages of ResNet in shallow detail extraction. In contrast, our approach achieves superior performance without introducing additional techniques, surpassing the combination of the two methods.

\begin{table}
  \centering
  \small
  \renewcommand\tabcolsep{2.7pt}
  \renewcommand{\arraystretch}{0.96}
  \caption{Comparison to state of the art KD methods with heterogeneous architectures on CIFAR-100. \underline{Underlined}: the second best result. \textbf{Bold}: the best result. }
  \begin{tabular}{cc cc cccc cccc>{\columncolor[gray]{0.85}}c}
    \toprule
    \multirow{2}{*}{Teacher} & \multirow{2}{*}{Student} & \multicolumn{2}{c}{From Scratch} & \multicolumn{4}{c}{Logits-based} & \multicolumn{5}{c}{Feature-based}                                                                                                                                             \\
    \cmidrule(lr){3-4} \cmidrule(lr){5-8} \cmidrule(lr){9-13}
                             &                          & T.                                & S.                              & KD                           & DKD     & DIST   & OFA   & FitNet               & CC                            & RKD                           & CRD              &  LFCC           \\
    \midrule
    \multicolumn{2}{c}{\emph{CNN-based students}}                                                                                                                                                                                                                                                            \\
    \midrule
    Swin-T                   & ResNet-18                 & 89.26                             & 74.01                           & 78.74             & 80.26                          & 77.75              & \underline{80.54}   & 78.87                          & 74.19 & 74.11 & 77.63     & \bf{83.56}        \\
    ViT-S                   & ResNet-18                 & 92.04                             & 74.01                       & 77.26                          & 78.10                          & 76.49        & \underline{80.15}                & 77.71 & 74.26 & 73.75  & 76.42 & \bf{81.45}        \\
    Mixer-B/16               & ResNet-18                 & 87.29                             & 74.01                           & 77.79 & 78.67                          & 76.36                       & \underline{79.39}           & 77.15                           & 74.26 & 73.75 & 76.42    & \bf{81.96}  \\
    Swin-T                   & MobileNetV2              & 89.26                             & 73.68                        & 74.68             & 71.07                          & 72.89              & \underline{80.98}           & 74.28                            & 71.19   & 69.00 & 79.80  & \bf{84.23}    \\
    ViT-S                   & MobileNetV2              & 92.04                             & 73.68                           & 72.77 & 69.80                          & 72.54                          & \underline{78.45}         & 73.54                            & 70.67 & 68.46 & 78.14  &\bf{82.06}      \\
    Mixer-B/16              & MobileNetV2              & 87.29                             & 73.68                          & 73.33 & 70.20                          & 73.26                        & \underline{78.78}            & 73.78 & 70.73 & 68.95 & 78.15    & \bf{82.49}    \\
    \toprule
    \multicolumn{2}{c}{\emph{Transformer-based students}}                                                                                                                                                                                                                                                            \\
    \midrule
    ConvNeXt-T               & DeiT-T                   & 88.41                             & 68.00                           & 72.99             & 74.60                & 73.55              & \underline{75.76}            & 60.78  & 68.01 & 69.79 & 65.94   & \bf{77.95}    \\
    Mixer-B/16               & DeiT-T                   & 87.29                             & 68.00                           & 71.36             & 73.44                & 71.67              & \underline{73.90}            & 71.05  & 68.13 & 69.89 & 65.35   & \bf{78.09}    \\
    ConvNeXt-T               & Swin-P                   & 88.41                             & 72.63                          & 76.44             & 76.80                 & 76.41          & \underline{78.32}          & 24.06 & 72.63 & 71.73 & 67.09   & \bf{79.21}    \\
    Mixer-B/16               & Swin-P                   & 87.29                             & 72.63                           & 75.93             & 76.39        & 75.85                      & \bf{78.93}       
 & 75.20 & 73.32 & 70.82 & 67.03    & \underline{78.23}       \\
    \toprule
    \multicolumn{2}{c}{\emph{MLP-based students}}                                                                                                                                                                                                                                                            \\
    \midrule
    ConvNeXt-T               & ResMLP-S12               & 88.41                             & 66.56                  & 72.25             & 73.22                   & 71.93              & \bf{81.22}             
          & 45.47 & 67.70 & 65.82 & 63.35 & \underline{79.08}      \\
    Swin-T                   & ResMLP-S12               & 89.26                             & 66.56                  & 71.89             & 72.82                   & 11.05              & \bf{80.63}           
          & 63.12 & 68.37 & 64.66 & 61.72 & \underline{78.88}      \\
    \bottomrule
  \end{tabular}
  \label{tab:cifar}
  \vspace{-0.3cm}
\end{table}

\subsection{Performance on CIFAR-100}

The experiments on the CIFAR-100 dataset are divided into three groups based on the family to which the student models belong, involving 12 heterogeneous teacher-student combinations. The proposed LFCC method is compared with 8 other state-of-the-art knowledge distillation methods. The results are presented in Table \ref{tab:cifar}.

LFCC significantly outperforms existing heterogeneous distillation methods when CNN-based model or DeiT-T is used as a student. Meanwhile, when MLP-based model is employed as a student, LFCC tends to yield suboptimal results. CIFAR-100, being a relatively small-scale dataset, logit-based methods typically outperform most feature-based methods, especially when the student model is belong to Transformer or MLP families. This indicates substantial disparities in feature representations among the models with heterogeneous architectures. Previous feature-based methods attempt to mimic the teacher's feature representations at local details, which may be detrimental, leading to overfitting on small datasets. LFCC extracts low-frequency components of features and aligns them in a compact space between teacher and student, uncovering commonalities between the models with heterogeneous architectures and significantly enhancing the performance and robustness of feature-based methods.

By comparing Table \ref{tab:imagenet} and Table \ref{tab:cifar}, we observe that the performance of distillation methods is influenced by the architectures of teacher and student models as well as the scale of the dataset. Despite our method being solely based on CNN architecture, it still achieves state-of-the-art results in most scenarios. This indicates that by extracting low-frequency components and compressing the feature alignment space, LFCC identifies commonalities in feature representations among models of various architectures. This exciting finding overturns the current situation where feature-based methods are deemed unsuitable for heterogeneous knowledge distillation.

\begin{figure}
  \centering
  \includegraphics[width=\linewidth]{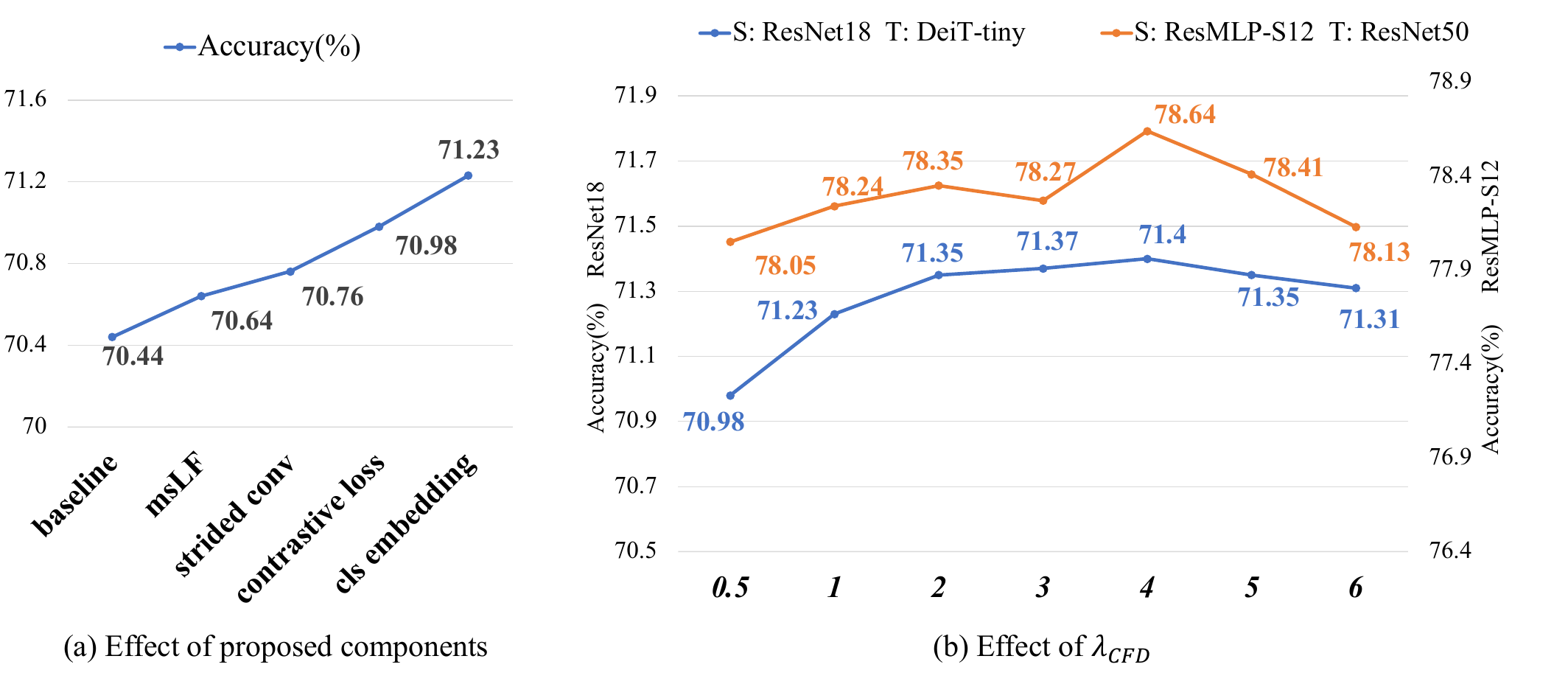}
  \caption{Ablation studies on ImageNet-1K. (a) Effect of proposed components. We incrementally incorporate the proposed components into the vanilla teacher-student framework to validate their efficacy. \(\lambda_{\text{CFD}}\) is set as 1.0 in all experiments. (b) We perform a comprehensive assessment of the impact of \(\lambda_{\text{CFD}}\) on two distinct teacher-student configurations. }
  \label{fig:abla}
  \vspace{-0.2cm}
\end{figure}

\subsection{Ablation study}
\label{sec:abla}

\paragraph{Effect of the proposed components.}

We systematically incorporate the proposed components into the vanilla teacher-student (ResNet18 vs DeiT-tiny) framework to comprehensively assess their efficacy. The findings are depicted in Figure \ref{fig:abla}(a). The results strongly validate the effectiveness of our methodology, with each component contributing to a marked improvement in the student model's performance. It is noteworthy that within our learnable low-pass filters, both the multi-scale low-pass filtering and convolutional downsample pathways positively impact the distillation process, highlighting their complementary functions.

\begin{wraptable}{t!}{7.0cm}
  \vspace{-0.5cm}
  \small
  \centering
  \renewcommand\tabcolsep{5.0pt}
  \renewcommand{\arraystretch}{0.9}
  \caption{Effect of number and position of aligned module on ImageNet-1K.}
  \vspace{-0.2cm}
  \begin{tabular}{cccc cc}
    \toprule
    \multicolumn{4}{c}{Stage} & \multicolumn{1}{|c}{T: DeiT-T}   & T: ResNet50 \\
    \cmidrule(lr){1-4}
       1  &  2  &  3  &  4    & \multicolumn{1}{|c}{S: ResNet18} & S: ResMLP-S12   \\
    \midrule
       -  &  -  &  -  &  -    & \multicolumn{1}{|c}{70.59}       & 77.32       \\
     \checkmark & - & - & -   & \multicolumn{1}{|c}{70.72}       & 77.41       \\
    - & \checkmark & - & -    & \multicolumn{1}{|c}{70.69}       & 77.49       \\
    - & - & \checkmark & -    & \multicolumn{1}{|c}{70.91}       & 77.67       \\
    - & - & - & \checkmark    & \multicolumn{1}{|c}{71.21}       & 77.94       \\
    \midrule
    - & - & \checkmark & \checkmark  & \multicolumn{1}{|c}{71.33}  & 78.32       \\
    - & \checkmark & \checkmark & \checkmark & \multicolumn{1}{|c}{71.36}       & 78.54       \\
    \checkmark & \checkmark & \checkmark & \checkmark & \multicolumn{1}{|c}{\bf{71.40}}  & \bf{78.64}  \\
    \bottomrule
  \end{tabular}
  \label{tab:ablation}
\end{wraptable}

\paragraph{Weight of contrastive feature loss.}

Within our LFCC framework, a diverse array of loss functions is employed, mandating the identification of an optimal equilibrium among these components.  Considering the extensive search for combinations of weights to be impractical, we fix the weights for cross-entropy and Kullback-Leibler divergence empirically, focusing solely on tuning \(\lambda_{\text{CFD}}\). Experiments are executed on two distinct teacher-student pairs: (ResNet18 vs DeiT-tiny) and (ResMLP-S12 vs ResNet50).  The results, as illustrated in Figure \ref{fig:abla}(b), indicate that the LFCC framework achieves peak performance at \(\lambda_{\text{CFD}}=4\).  Consequently, \(\lambda_{\text{CFD}}=4\) is adopted as the default setting for subsequent experiments.

\paragraph{Position and number of aligned modules.}

Our LFCC inserts auxiliary feature alignment modules into intermediate layers to capture the intercommunity between teacher and student. In our experiments, we investigate the optimal positioning and quantity of alignment modules. Both the student and teacher models are divided into four stages, as represented by corresponding numbers in Table \ref{tab:ablation}. We observe that while alignment modules at each stage contribute to the training of the student model, inserting an alignment module into the final stage yields the most significant benefits, particularly for ResNet18. Experimental results indicate that incorporating alignment modules in all stages provides the most pronounced enhancement to the student's performance.

\begin{table}
  \small
  \centering
  \renewcommand\tabcolsep{4.2pt}
  \renewcommand{\arraystretch}{0.9}
  \caption{KD methods with homogeneous architectures on ImageNet-1K. \emph{T}: ResNet34, \emph{S}: ResNet18.}
  \begin{tabular}{c|cc|ccccccc>{\columncolor[gray]{0.85}}c}
    \toprule
             & T.    & S.    & KD~\cite{distilling2015hinton} & OFD~\cite{heo2019comprehensive} & Rev~\cite{chen2021distilling} & CRD~\cite{contrastive2019tian} & DKD~\cite{decoupled2022zhao} & DIST~\cite{knowledge2022huang} & OFA~\cite{ofa2023hao}  & LFCC \\
    \midrule
    Acc & 73.31 & 69.75 & 70.66 & 70.81 & 71.61 & 71.17 & 71.70 & 72.07 & 72.10 &  \textbf{72.26} \\
    \bottomrule
  \end{tabular}
  \label{tab:homo}
  \vspace{-0.2cm}
\end{table}

\begin{wrapfigure}{r}{0.5\textwidth}
  \begin{center}
    \includegraphics[width=0.48\textwidth]{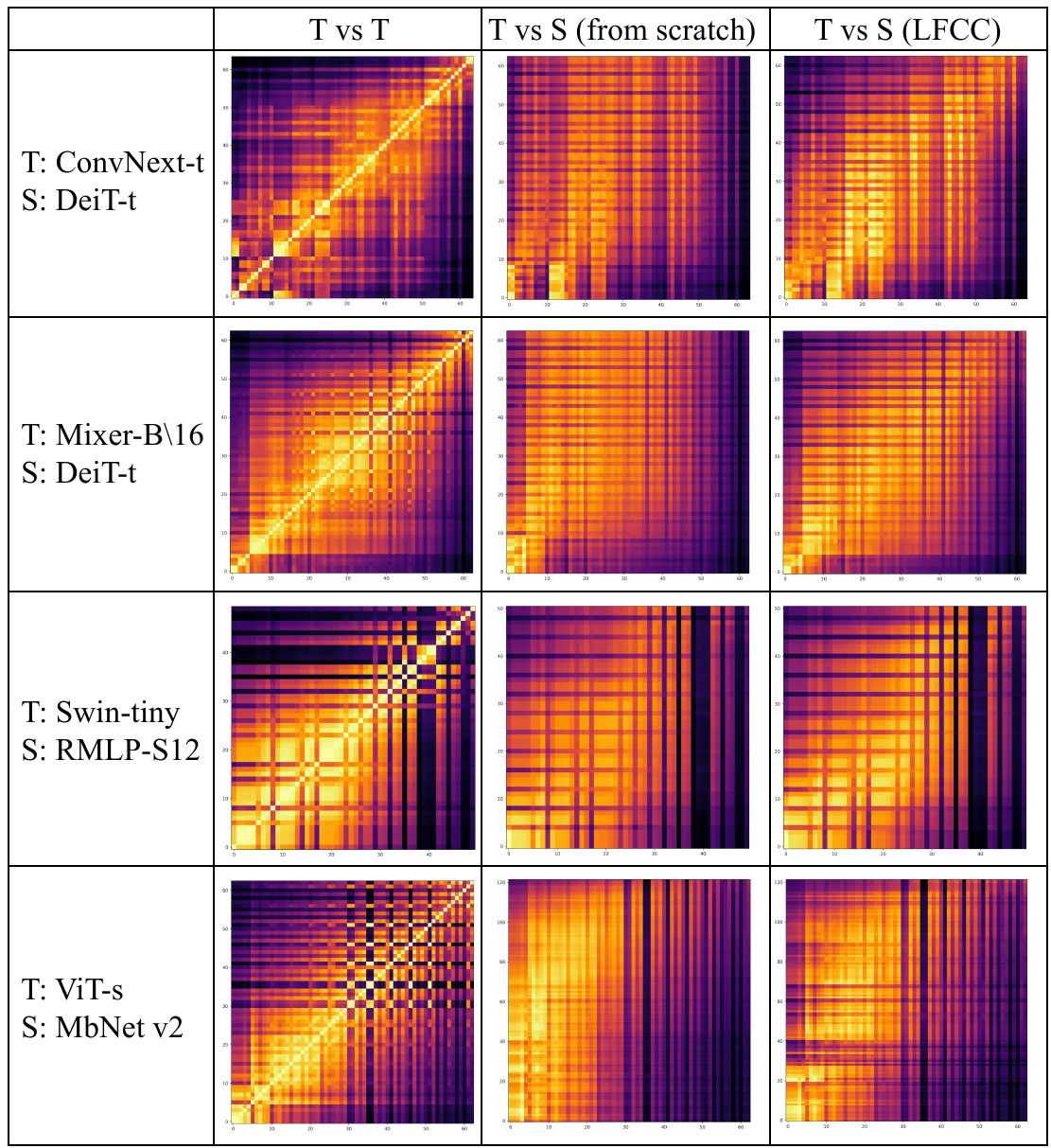}
  \end{center}
  \caption{CKA-assessed similarity heatmap of intermediate features, where RMLP-S12 denotes ResMLP-S12 and MbNet represents MobileNet V2. Each row of the figure representing the feature similarity for a given model in comparison to the teacher model across three conditions: the teacher model itself, a student model trained from scratch, and a student model trained via LFCC.}
  \label{fig:vis}
  \vspace{-0.2cm}
  
\end{wrapfigure}

\paragraph{Distillation in homogeneous architectures.}

We further validate the performance of our proposed Low-Frequency Components-based Contrastive Knowledge Distillation (LFCC) on the benchmark for homogenous architecture distillation, which is widely recognized. The results, as presented in Table \ref{tab:homo}, indicate that LFCC surpasses other baseline distillation approaches when applied to the teacher-student pair (ResNet18 vs ResNet34), thereby convincingly highlighting its robust capability and generalizability.

\paragraph{Visualization of the feature alignment.}

Centered Kernel Alignment (CKA)\cite{cortes2012algorithms, kornblith2019similarity} serves as a metric for assessing feature similarity, enabling cross-architectural comparisons by accommodating inputs of different dimensions. We qualitatively evaluate the efficacy of the proposed LFCC in heterogeneous architectural feature alignment via CKA. The comparison is conducted on four teacher-student pairs, assessing their intermediate features both before and after guidance from a heterogeneous teacher. The results on the CIFAR-100 dataset are depicted in Figure \ref{fig:vis}. Within the LFCC framework, convergence of the intermediate features between the student and teacher models is observed, signifying the efficacy of our method in establishing feature alignment across heterogeneous architectures and consequently improving the student model's performance.

\section{Conclusion}

To mitigate the diminished performance of feature-based distillation methods across heterogeneous architectures, we propose an innovative framework, named Low-Frequency Components-based Contrastive Knowledge Distillation (LFCC) .  This framework is engineered to discover shared low-frequency components within the intermediate features of teacher and student models, aligning them within a condensed space. Consequently, LFCC neutralizes the intrinsic disparities in feature representation inherent to heterogeneous models, facilitating the precise conveyance of knowledge encapsulated in the teacher's intermediate features. Furthermore, capitalizing on the innate pairing within the teacher-student construct, we construct batches of teacher and student features into pairs of positive and negative samples, thereby engaging in sample-level contrastive learning.  The efficacy of our approach is corroborated by its robust performance on the formidable benchmarks of ImageNet-1K and CIFAR-100.

{\small
\bibliographystyle{unsrt}
\bibliography{ref}
}

\end{document}